# Assessment, criticism and improvement of imprecise subjective probabilities for a medical expert system

David J Spiegelhalter*
Rodney C G Franklin+
Kate Bull+

* MRC Biostatistics Unit, 5 Shaftesbury Rd, Cambridge, CB2 2BW, UK.
+ Hospital for Sick Children, Great Ormond St, London, WC1 3JH, UK.

## Abstract

Three paediatric cardiologists assessed nearly 1000 imprecise subjective conditional probabilities for a simple belief network representing congenital heart disease, and the quality of the assessments has been measured using prospective data on 200 babies. Quality has been assessed by a Brier scoring rule, which decomposes into terms measuring lack of discrimination and reliability. The results are displayed for each of 27 diseases and 24 questions, and generally the assessments are reliable although there was a tendency for the probabilities to be too extreme. The imprecision allows the judgements to be converted to implicit samples, and by combining with the observed data the probabilities naturally adapt with experience. This appears to be a practical procedure even for reasonably large expert systems.

## 1. Introduction

A belief-network representation of medical knowledge consists of two components (Pearl, 1988; Lauritzen and Spiegelhalter, 1988); a qualitative acyclic directed graph expressing conditional independence assumptions concerning the variables of interest, and a quantitative component comprising the set of conditional probability tables appropriate to the graph.

In this paper we assume the qualitative graph is given and fixed, and concentrate on the problem of assessment, criticism and refinement of the conditional probabilities in the light of the data. In considering the practical aspects of implementing expert systems based on belief networks, it has often been said that a major difficulty is obtaining the precise quantities that are apparently required. We shall point out that the opposite holds, and that the true power of a probabilistic representation is its ability not only to deal with <u>imprecise</u> probability assessments, but to welcome them as providing a natural basis for the system to improve with experience.

The techniques will be illustrated by a continuing study aimed at developing a system that represents knowledge concerning the presentation of congenital heart disease in newborn infants. A brief background is given in the next section, followed by a description of the procedures and results from the first attempt at eliciting imprecise subjective probabilities. A prospective set of data is available to contrast with the assessments, and the idea of 'proper scoring rules' is introduced and a simple example shown. A technique for turning the imprecise assessments into implicit samples is illustrated and it is shown how this leads to a natural means of revising the initial probability judgement with the data. The discussion focuses on how the expressed imprecision might be taken into account when criticising the assessments.

## 2. Background, Assessments and Data

Each year about 200 new-born infants with suspected congenital heart disease are referred to Great Ormond Street Hospital (GOS), which acts as one of the regional centres for the South-East of England. One of the most important prognostic factors is the health of the baby upon arrival at the hospital, and hence it is vital that appropriate treatment starts before transportation from the referring hospital. Therefore a project exists to provide a system to help the GOS clinician with making a diagnosis over the telephone, based on information provided by the referring paediatrician. The current version of the system is strictly algorithmic (Franklin et al., 1989) and is now being evaluated.

As a second stage to the project it is intended to construct a system based on a belief network. This will explicitly represent the mechanism by which a pathological anomaly in the heart may lead to specific pathophysiological disturbances, which in turn tends to lead to abnormal feature that may or may not be identified and reported over the telephone. The uncertainty attached to each link in this network needs to be expressed as a conditional probability table.

There are 27 main diseases under consideration, and 24 questions asked over the telephone, each of which may have 2 to 5 possible responses. As a preliminary study into the feasibility of eliciting subjective probabilities, three paediatric cardiologists assessed for each disease, assumed mutually exclusive in this exercise, the chance of a new-born presenting with each possible finding. We emphasise that the clinicians had in mind features elicited at the specialist hospital, rather than as reported over the telephone. We also note that such assessments will need to be extensively supplemented to provide the quantitative component of the full belief network. The clinicians were asked to avoid unjustifiably precise probability assessments, and instead to give a consensus range for each conditional probability. However the precise interpretation of the range was not specified, and no technique for assessing the range was recommended. A small subset of the assessments obtained in this pilot study are shown in Table 1.

|  |  | Disease | | |
|---|---|---|---|---|
| Question: | Feature: | Non-urgent heart disease | Aortic stenosis | Hypoplastic left heart |
| Main problem? | Cyanosis | 0-0% | 0-0% | 5-10% |
|  | Heart failure | 0-0% | 90-95% | 90-95% |
|  | Asymptomatic murmur | 100-100% | 2-7% | 1-2% |
|  | Arrythmia | 0-0% | 0-0% | 0-0% |
|  | Other | 0-0% | 0-0% | 0-0% |
| Grunting? | Yes | 5-10% | 5-15% | 30-40% |
|  | No | 90-95% | 85-95% | 60-70% |

Table 1. Judgements (given as percentage ranges) of proportions of newborns which will present with particular features, for each of three diseases. For example, the three experts thought that between 30 and 40% of babies with hypoplastic left heart would exhibit the feature of 'grunting'.

We note that the clinicians were reasonably precise in their judgements, and were quite categorical in ruling out particular 'main problems' for each of the diseases. If we consider the above segment as comprising 15 independent assessments, (since for each question one of the responses must hold) there were a total of 37 x 27 = 999 independent assessments made, for questions on, for example, clinical signs, heart rate, pulses, chest X-ray, electro-cardiogram, and blood gases.

The data available comprise 200 cases collected prospectively, after the probability assessments had been made, using a standardised form on which the features described over the telephone were recorded (with some missing data). Definitive diagnosis was made by echo-cardiogram or cardiac catheterisation after transportation to Great Ormond Street. The results corresponding to Table 1 are given in Table 2.

|              |                      | Disease                      |                    |                        |
|--------------|----------------------|------------------------------|--------------------|------------------------|
| Question:    | Feature:             | Non-urgent heart disease     | Aortic stenosis    | Hypoplastic left heart |
| Main problem? | Cyanosis            | 2 ( 10%)                     | 1 ( 25%)           | 2 ( 10%)               |
|              | Heart failure        | 5 ( 24%)                     | 2 ( 50%)           | 18 ( 90%)              |
|              | Asymptomatic murmur  | 11 ( 52%)                    | 1 ( 25%)           | 0 ( 0%)                |
|              | Arrythmia            | 0 ( 0%)                      | 0 ( 0%)            | 0 ( 0%)                |
|              | Other                | 3 ( 14%)                     | 0 ( 0%)            | 0 ( 0%)                |
| Grunting?    | Yes                  | 0 ( 0%)                      | 0 ( 0%)            | 7 ( 37%)               |
|              | No                   | 21 (100%)                    | 4 (100%)           | 12 ( 63%)              |

Table 2. Data recorded on prospectively collected cases, corresponding to assessments shown in Table 1.

Two conclusions are immediately apparent on comparing the expert judgements of Table 1 with the data in Table 2. The good news is that the judgments concerning hypoplastic left heart are extraordinarily good. The bad news is that the experts' opinion about what cases of non-urgent heart disease have as a main problem is sadly out of line with what is reported over the telephone. Formal means of measuring such discrepancies are now considered.

### 3. Criticising the probability assessments

There is substantial literature on the criticism of subjective probabilities stemming largely from work on evaluating probabilistic precipitation forecasts made by US weather forecasters (see, for example, Dawid (1986) for a review, also Murphy and Winkler (1984)). Apparently all this work has focused on evaluating precise probability forecasts. In the simplest case, suppose a probability vector $\underline{p} = (p_1, p_2, ... p_k)$ has been given to a question E which can take on one of k values; after response j is observed we have a vector $\underline{e} = (e_1, ..., e_k)$, where $e_j = 1$, $e_i = 0$ for $i \neq j$. Then a 'scoring rule' is some function $S(\underline{e}, \underline{p})$ that penalises the assessor if a low probability has been assigned to an event that occurs. In weather forecasting the 'Brier' score

$$S(\underline{e}, \underline{p}) = \tfrac{1}{2} \Sigma (e_i - p_i)^2$$

is most common, where the multiplier $\tfrac{1}{2}$ ensures a range of 0 to 1. We note that the Brier score can also be expressed as $\tfrac{1}{2}(2 - p_r + \Sigma p^2_i)$, where $p_r$ is the probability given to the event that actually occurred. Another common rule is the 'logarithmic' score

$$S(\underline{e}, \underline{p}) = -\log(p_r)$$

which has been used by Shapiro (1977); essentially this measures the assessed log likelihood of the events that occurred, and is extremely punishing to widely wrong predictions. These two rules are both 'strictly proper', in the sense that an assessor's expected penalty is minimised by stating his 'true' probability. The absolute deviation score $S(\underline{e}, \underline{p}) = \Sigma |e_i - p_i|$ is <u>not</u> proper, although it has been used by Zagoria and Reggia (1983); if when forecasting a binary variable an observer is to be assessed by this rule, it will pay him or her to quote 'p = 1' whenever their true belief is $> \tfrac{1}{2}$, and 'p=0' otherwise ( if their true belief is $\tfrac{1}{2}$ then they could quote any value for p as they will always expect to score $\tfrac{1}{2}$).

In this paper we shall turn the imprecise assessments into precise assessments to be judged by the Brier scoring rule. This apparent loss of information can be well justified when it comes to processing a single case. Spiegelhalter and Lauritzen (1989) have shown that if the conditional probabilities in a belief network are themselves considered as imprecise quantities with their own attached uncertainty, then in processing a case this imprecision should be ignored and a single 'mean' probability value used in order to derive appropriate evidence propagation on the case in hand. (We emphasise, however, that the evidence extracted from that case will in turn revise the conditional probability to be adopted in processing the next case). Furthermore, suppose we had some distribution over our subjective probability $p_i$ for an event $e_i$ and had to choose a representative value $P_i$ knowing it was to be criticised according to the Brier score. Then it is straightforward to show that our expected contribution to the overall score for that question is minimised if we choose $P_i$ to be the mean of our distribution.

We therefore consider only the adequacy of the <u>midpoints</u> of the intervals provided for each of the events, although later we shall discuss how the imprecision becomes important if we allow the assessments to adapt as data accumulates.

If we consider Table 1, we see, for example, that the 'precise' probability vector $\underline{p}$ for the question 'grunting?' in aortic stenosis is simply (0.10, 0.90). However, a problem arises if the experts have given intervals for a set of responses to a question whose midpoints do not add to 1, as, for example, is the case for 'main problem?' in aortic stenosis and hypoplastic left heart (and in many other instances not shown here). In this situation the midpoints of the intervals have been simply rescaled to add to 1; for example, 'main problem?' in aortic stenosis has unadjusted probability vector (0, 0.925, 0.045, 0, 0), which is rescaled to (0, 0.954, 0.046, 0, 0).

We now consider the calculation of the Brier score, and use the single case of aortic stenosis presenting with asymptomatic murmur as an example. We have $\underline{p} = (0, 0.954, 0.046, 0, 0)$, $\underline{e} = (0, 0, 1, 0, 0)$ and hence $B = \frac{1}{2} (0 + 0.954^2 + (1-0.046)^2 + 0 + 0) = 0.9101$ for that question.

These scores may be analyzed in many different ways. The mean score, on a total of 3944 questions asked, was 0.12, while the mean score within disease varied from 0.08 to 0.22. However a disease may score highly simply because it does not present in a clearcut way and hence many probabilities near 0.5 are given to events, which will inevitably lead to a poor score. In order to measure the quality of the probability assessments, as opposed to the distinguishability of the disease, we clearly have to analyze the score more carefully.

### 4. Discrimination and Reliability

There has been extensive work on decompositions of scoring rules into terms that reflect different aspects of an assessor's skill, see, for example, Yates (1982) and DeGroot and Fienberg (1983). Here we shall consider simple decompositions into a term that expresses 'lack of discrimination' (how far away from 0 and 100 the assessments are) and 'lack of reliability' (how untrustworthy the judgements are). Good reliability means, for example, that if an assessor provides a probability of .7 for a response, the about 70% of the questions will result in that response. Reliability appears more important than discrimination in the context of describing accurately the presentation of a disease.

Hilden et al (1978) provide a general method of obtaining a decomposition. Under the hypothesis that our experts provide perfectly reliable judgements, then the Brier score for a single question on a single case has expectation.

$$E_0(B) = \tfrac{1}{2} E_0[\Sigma(e_i - p_i)^2] = \tfrac{1}{2} E_0[(1 - 2p_r + \Sigma p^2_i)] = \tfrac{1}{2}(1 - \Sigma p^2_i)$$

since $E_0(p_r) = \Sigma p^2_i$. Deviations of B from this figure measure lack of reliability, and we denote as R the statistic

$$R = B - E_0(B) = \tfrac{1}{2}[1 - 2p_r + \Sigma p^2_i - 1 + \Sigma p^2_i] = \Sigma p^2_i - p_r.$$

The mean of R will be positive if the assessor tends to make too extreme probability judgements (over-confidence), and be negative if his assessments are not extreme enough (diffidence).

Figures 1 and 2 show how the mean discrimination and reliability vary between the 27 diseases and the 24 questions that are asked. Disease assessments showing poor discrimination and good reliability, such as PFC (persistent foetal circulation), are characterised by having a varied presentation (probabilities not near 0 and 100) which is well-understood by the experts. Diseases such as AVSD+ (AV septal defect + outflow obstruction) and NUHD (non-urgent heart disease) are perceived as having a very consistent presentation (good discrimination) but this perception seems to be at fault.

Questions such as 'heart failure?' display poor discrimination and excellent reliability, in that there is a wide variation in its incidence among diseases, but these incidences are known reasonably well.

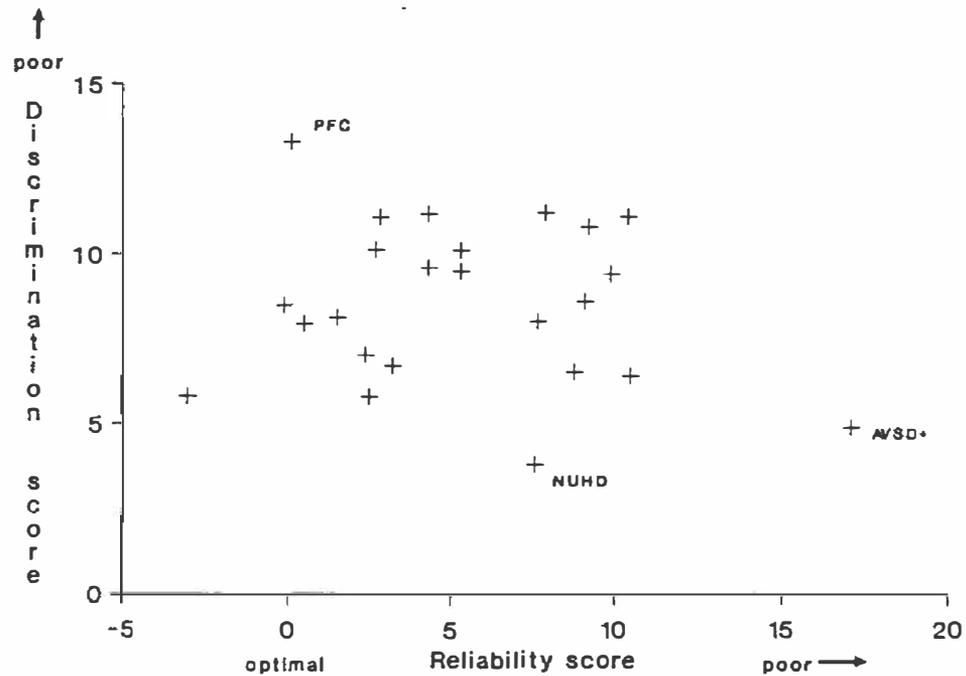

Figure 1. Mean discrimination and reliability scores of probability assessments, when grouped according to disease (see text for explanation of abbreviations).

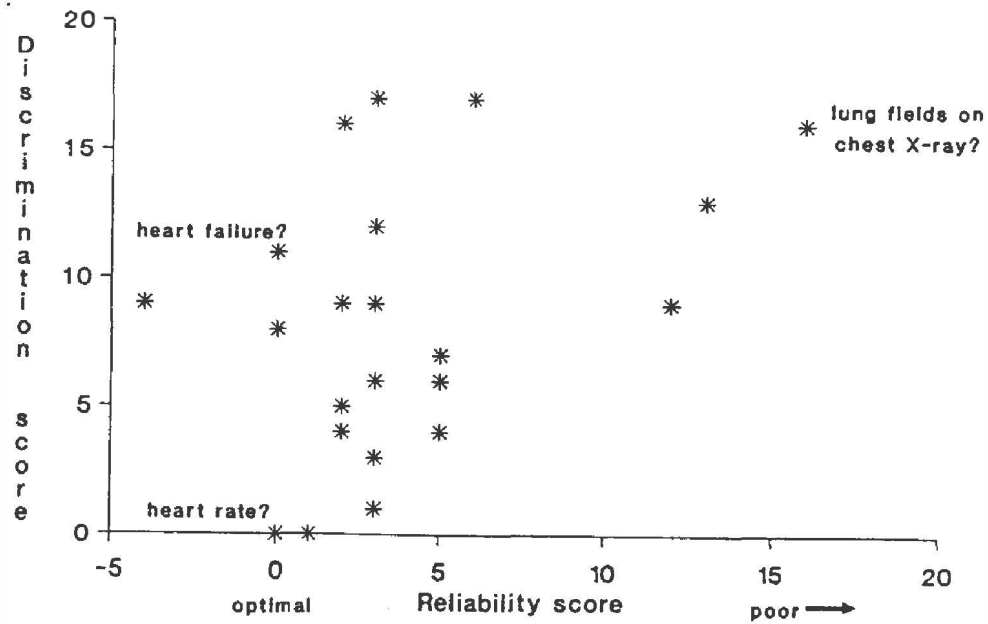

Figure 2  Mean discrimination and reliability scores for probability assessments, when grouped according to question.

'Heart rate?' is a question with a categorical response in each disease, which is correctly known. However, 'lung fields on chest X-ray?' shows wide variation in response within diseases, and these variations in presentation are not well-estimated by the experts. Plots such as Figures 1 and 2 identify where further questioning of the experts may be necessary.

A graphical representation of reliability is valuable and a number of suggestions have been made (Hilden et al, 1978). The simplest method is to group the probability assessments $p_i$, for all responses to all questions in all diseases, into a small set of categories $x_1,...,x_k$. If assessments are assumed only to be made at these discrete values, then we can count the total number $n_k$ of times a probability $x_k$ has been given to a prospective event, and the fraction $f_k$ of $n_k$ in which these events occurred. Figure 3 shows a plot of $f_k$ against $x_k$, where assessments are considered in twelve groups; 0%, 1-10%, 11-20%,..., 90-99%, 100%. This brings home the striking reliability of the assessors. One problem is that out of 3210 events that were given probability zero, 74 actually occurred. A possible solution to this problem is given in the discussion.

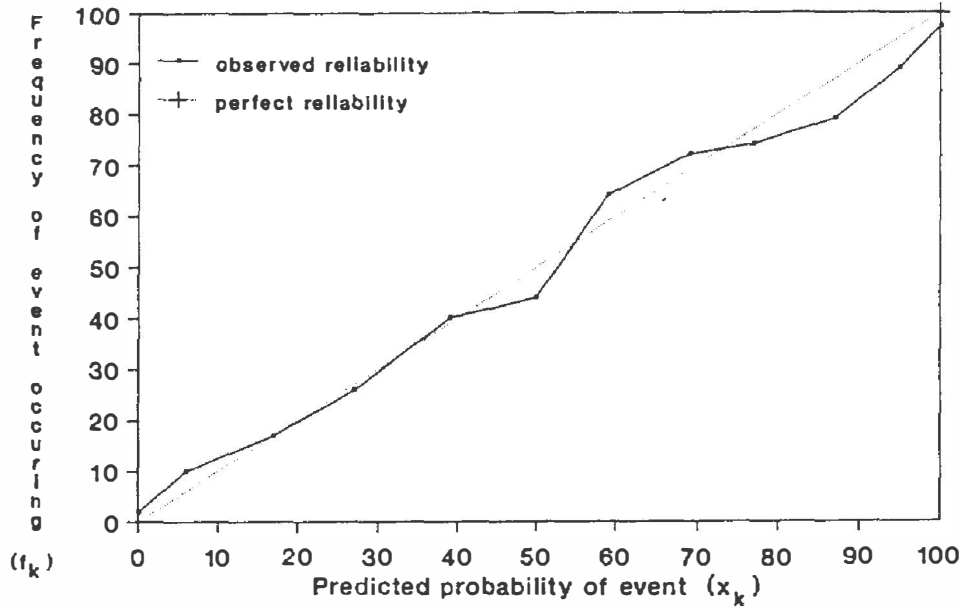

Figure 3  Overall reliability of subjective probability assessments; for example, of 297 events given probability between 41% and 50%, 131 (44%) actually occurred. (Midpoints of original consensus probability intervals have been used).

## 5. Learning from experience

As mentioned in the introduction, the imprecision expressed in the ranges for the probability assessments is crucial in providing a mechanism for the experts' judgements (however good they may appear to be) to be tempered by experience. Essentially we assume there is some 'true' probability vector $\underline{p}$ appropriate to a particular question asked of a patient with a particular disease. The experts provide a probability distribution $f(\underline{p})$ for the vector, and when data D is observed, the opinion about $\underline{p}$ is revised by Bayes theorem to be $f(\underline{p}|D) \propto f(D|\underline{p}) f(\underline{p})$. Spiegelhalter and Lauritzen (1989) discuss a number of different models for $f(\underline{p})$, and we only consider the simplest here. It is well known that if $f(\underline{p})$ is assumed to be a Dirichlet distribution (a beta distribution where questions have only two responses) then the prior opinion is equivalent to an 'imaginary' sample of cases that represents the experts' experience ; Bayes theorem simply updates that imaginary sample with real cases. In effect the conditional probabilities are stored as fractions of whole numbers, rather than single numbers between 0 and 1.

Let us illustrate using the assessments in Table 1 for 'grunting?' in hypoplastic left heart. The expert assessment of 30-40% can be thought of as summarising a distribution, and we shall, somewhat arbitrarily, assume that there is about a 2 to 1 chance that the true frequency lies in that interval. This interpretation is equivalent to assuming the interval is a one standard error interval calculated after observing n cases of the disease of which a fraction 0.35 displayed grunting, where

$$(0.30, 0.40) = 0.35 \pm 0.35(1-0.35)/n$$

using standard binomial statistical theory. Hence $n = 0.35 \times 0.65/0.05^2 = 91$. Thus we take our experts' opinion as equivalent to having observed 32 instances of 'grunting' among 91 cases of hypoplastic left heart (HLH).

The probability $32/91 = 0.352$ will be used for the next case, but if this happens to be HLH with grunting, the fraction will increase to 33 out of 92. The probability $33/92 = 0.359$ will then be used for the next case. We note that adaptation may be slow when the experts are initially reasonably confident, and from Table 2 we see that after a year's worth of 200 cases the fraction has only shifted to $(32+7)/(91+19) = 0.355$, reflecting the accurate initial assessment. If, in contrast, the experts acknowledge their uncertainty about the probability and provide a wide initial range, adaptation will be considerably faster.

A certain arbitrariness exists in dealing with more than two responses in which some assessments are more precise than others. Currently we use the formula

$$n = \text{midpoint} \times (1-\text{midpoint})/(\text{half the range of the interval})^2$$

for each response, and adopt the lowest n as the implicit sample size underlying the expert judgment, i.e. we fix on the most imprecise assessment.

Table 3 shows the results of learning about the aortic stenosis judgements shown in Table 1.

| Question: | Feature: | Initial Judgements | Initial probs | Implicit sample size | Observed sample | Combined samples & probs | |
|---|---|---|---|---|---|---|---|
| Main problem? | Cyanosis | 0-0% | 0.0 | 0.0 | 1 | 1.0 | 0.014 |
| | Heart failure? | 90-95% | 0.945 | 65.8 | 2 | 67.8 | 0.929 |
| | Asymptomatic murmur | 2-7% | 0.046 | 3.2 | 1 | 4.2 | 0.058 |
| | Arrythmia | 0-0% | 0.0 | 0.0 | 0 | 0.0 | 0.0 |
| | Other | 0-0% | 0.0 | 0.0 | 0 | 0.0 | 0.0 |
| | | | 1.00 | 69.0 | 4 | 73.0 | 1.000 |
| Grunting? | Yes | 5-15% | 0.10 | 3.6 | 0 | 3.6 | 0.090 |
| | No | 85-95% | 0.90 | 32.4 | 4 | 36.4 | 0.910 |
| | | | 1.00 | 36.0 | 4 | 40.0 | 1.000 |

Table 3. Initial judgements transformed to point probabilities, and then to implicit samples. When combined with the observed data the revised conditional probabilities are obtained.

We note fractional implicit samples are quite reasonable. Once again, the major problem is the confidence expressed by 0-0% probability ranges. If, as in Table 3, other options are given less extreme judgements, then this procedure will revise an initial 0-0% interval. However, if all possible answers to a question are given 0-0% or 100-100% assessments, then the implicit sample size is, strictly speaking, infinity and no learning can take place. We are currently experimenting with a range of 'large' sample sizes to use in place of infinity.

### 6. Discussion

We have briefly described a procedure for criticising and improving upon imprecise subjective probability assessments. Many issues remain open for further investigation. From a practical viewpoint, the pilot study has shown that reliable probability assessments can be obtained from experts although there is a tendency to them to be too extreme in their judgements. Since the observations have been obtained over the telephone we might expect additional variation to be presented which could explain this over-confidence, and our next task is to contrast these results with data obtain at the specialist hospital. Our results contrast with those of Leaper et al (1972) who criticised the use of clinicians' probability assessments for a simple network representing acute abdominal pain, although they did not explicitly identify the kind of errors made by the clinicians and only compared final diagnostic performance. One explanation is the different clinical domain, in that compared with acute abdominal pain the precise presentation of babies with congenital heart disease has been carefully studied.

From a technical perspective we need to investigate appropriate means of dealing with 'zero' assessments (currently our opinion is that they all should be treated as 0-4% on the basis of our empirical discovery of a 2% error rate). We are also examining alternative means of transforming intervals to implicit samples, so that the adequacy of our 'one standard error interval' assumption can be tested.

Strictly, the correct means of scoring these imprecise assessments is to introduce the data in single cases, updating probabilities sequentially, and use the revised judgements for each new case. This needs to be investigated for a number of different scoring rules, and tools developed for early automatic identification of poor initial assessments. This 'prequential' approach (predictive sequential) has been pioneered by Dawid (1984), and promises to form the basis for automatic monitoring of the performance of predictive expert systems.

## Acknowledgments

We are grateful to Philip Dawid for valuable discussions, and to the British Heart Foundation and the Science and Engineering Research Council for support.

## References

Dawid A P (1984). Statistical theory - the prequential approach. J. R. Stat. Soc. A, 147, 277-305.
Dawid A P (1986). Probability forecasting, Encyclopedia of Statistical Sciences, Vol 7, (eds Kotz and Johnson), J Wiley: New York, pp. 210-218.
DeGroot M H & Fienberg S E (1983). The comparison and evaluation of forecasters. Statistician, 32, 12-22.
Franklin R C G, Spiegelhalter D J, MaCartney F and Bull K (1989). Combining clinical judgements and statistical data in expert systems: over the telephone management decisions for critical congenital heart disease in the first month of life. Intl. J. Clinical Monitoring and Computing, 6 (to appear).
Hilden J, Habbema J D F & Bjerregaard B (1978). The measurement of performance in probabilistic diagnosis, III - methods based on continuous functions of the diagnostic probabilities. Methods of Information in Medicine, 17, 238-346.
Lauritzen S L & Spiegelhalter D J (1988). Local computation with probabilities on graphical structures, and their application to expert systems (with discussion). J. Roy. Stat. Soc. B, 50, 157-244.
Leaper D J, Horrocks J C, Staniland J R and de Dombal F T (1972) Computer-assisted diagnosis of abdominal pain using 'estimates' provided by clinicians. British Medical Journal, 4, 350-354.
Murphy A H & Winkler R L (1977). Reliability of subjective probability forecasts of precipitation and temperature. Applied Statistics, 26, 41-47.
Murphy A H & Winkler R L (1984). Probability forecasting in meteorology, J. Amer. Statist. Assoc., 79, 489-500.
Pearl J (1988). Probabilistic reasoning in intelligent systems. Networks of Plausible Inference. Morgan Kaufmann, California.
Shapiro A R (1977). The evaluation of clinical predictions. A method and initial application, New England Journal of Medicine, 296, 1509-1514.
Spiegelhalter D J and Lauritzen S L (1989). Sequential updating of conditional probabilities on directed graphical structures. Networks, (to appear).
Yates J F (1982). External correspondence: Decomposition of the mean probability score. Organisational Behaviour and Human Performance, 30, 132-156.
Zagoria R J & Reggia J A (1983). Transferability of medical decision support systems based on Bayesian classification, Medical Decision Making, 3, 501-510.